\documentclass[runningheads]{llncs}
\usepackage{graphicx}
\usepackage{multirow}
\usepackage{booktabs}
\usepackage{amsmath}
\usepackage{amssymb}

\usepackage{algorithm}
\usepackage{algorithmic}

\usepackage{cite}
\usepackage{url}
\usepackage[marginal]{footmisc}

\begin{document}
\title{Self-guided Few-shot Semantic Segmentation for Remote Sensing Imagery Based on Large Vision Models}
\titlerunning{Self-guided Large Vision Model (Few-shot SLVM)}
\renewcommand{\thefootnote}{}
%
\author{Xiyu Qi\inst{1,2,4,*} \and
Yifan Wu\inst{1,3,4,*} \and
Yongqiang Mao\inst{1,2,4} \and
Wenhui Zhang\inst{1,2,4} \and
Yidan Zhang\inst{1,2} }
\authorrunning{Qi. Author et al.}
%
\institute{Aerospace Information Research Institute, Chinese Academy of Sciences, Beijing 100190, China \and
Key Laboratory of Network Information System Technology (NIST), Aerospace Information Research Institute, Chinese Academy of Sciences, Beijing 100190, China \and
Key Laboratory of Technology in Geo-Spatial Information Processing and Application System, Aerospace Information Research Institute, Chinese Academy of Sciences, Beijing 100190, China \and
School of Electronic, Electrical and
Communication Engineering, University of Chinese Academy of Sciences, Beijing 100190, China}
\maketitle              
\footnote{*These authors contributed equally to this work and should be considered co-first authors.}
\begin{abstract}
The Segment Anything Model (SAM) exhibits remarkable versatility and zero-shot learning abilities, owing largely to its extensive training data (SA-1B). Recognizing SAM's dependency on manual guidance given its category-agnostic nature, we identified unexplored potential within few-shot semantic segmentation tasks for remote sensing imagery. This research introduces a structured framework designed for the automation of few-shot semantic segmentation. It utilizes the SAM model and facilitates a more efficient generation of semantically discernible segmentation outcomes. Central to our methodology is a novel automatic prompt learning approach, leveraging prior guided mask to produce coarse pixel-wise prompts for SAM. Extensive experiments on the DLRSD datasets underlines the superiority of our approach, outperforming other available few-shot methodologies.

\keywords{Remote sensing images \and visual foundation model \and semantic segmentation \and prompt learning.}
\end{abstract}
\section{Introduction}
Recent advancements in remote sensing technologies \cite{moselhi2020automated,mao2022beyond} have revolutionized the way we collect and analyze data relevant to our understanding of the Earth's surface. The data collected from these methods serve central to a multitude of applications such as weather forecasting \cite{gherboudj2016assessment}, environmental monitoring \cite{li2020review}, urban planning \cite{wellmann2020remote,mao2023elevation}, and defense intelligence \cite{michael2006realized}. Despite the wealth of data made available through remote sensing, effective utilization of this information poses a significant challenge. One bottleneck lies in the semantic segmentation of the remote sensing imagery \cite{qi2023pics,yi2019semantic}. Traditional methods rely heavily on manual guidance, which is both time-inefficient and susceptible to human error.\\
\indent The Segment Anything Model (SAM) \cite{kirillov2023segment}, an example of large vision models, demonstrates considerable versatility and zero-shot learning \cite{romera2015embarrassingly} abilities, powered by its rich training data. However, its ability to perform one-shot or few-shot semantic segmentation tasks for remote sensing imagery \cite{chen2022semi,wang2021dmml} remains largely underexplored. The SAM model’s design is category-agnostic, which inevitably forces reliance on manual guidance. Recognizing the potential and the need to optimize this aspect, we ventured into the integration of few-shot learning in this regard.\\
\indent In this research, we introduce a novel few-shot semantic segmentation algorithm designed to automate the semantic segmentation process termed Self-guided Large Vision Model (Few-shot SLVM). Our approach enables the use of the SAM model with the intent to achieve efficient generation of semantically rich segmentation outcomes. The cornerstone of this method is an innovative automatic prompt learning technique that leverages prior guided masks to produce coarse pixel-wise prompts for SAM, bypassing the need for intensive manual guidance.\\
\indent The framework we propose for few-shot semantic segmentation provides a promising avenue for the efficient parsing of remote sensing imagery. The system's capacity to produce high-quality segmented images with limited supervision is bound to drive advancements in various applications dependent on remote sensing. We focused our experiments on the DLRSD \cite{shao2018benchmark} datasets with the goal of validating the superiority of this approach over other existing methodologies for few-shot semantic segmentation, particularly in the context of remote sensing. \\
\indent To summarize, our major contributions are:
\begin{itemize}
    \item [1)]
    We introduce the Self-guided Large Vision Model (Few-shot SLVM), a novel few-shot semantic segmentation framework, that significantly automates the segmentation process for remote sensing imagery without heavy reliance on manual guidance.
    \item [2)]
    We propose an innovative 'automatic prompt learning' technique using the Segment Anything Model (SAM) for rendering coarse pixel-wise prompts, bringing a novel solution to semantic segmentation of remote sensing imagery.
    \item [3)]
    We carry out extensive benchmarking on the DLRSD datasets, showcasing the superiority of our methodology against existing few-shot segmentation techniques within the domain of remote sensing.
\end{itemize}

\section{Related Work}
\indent In this section, we initially introduce recent work in the field of semantic segmentation, laying the foundation for essential context with regards to our proposed method. Subsequently, we steer our focus towards the few-shot semantic segmentation techniques that underpin our proposed method while concluding this section with a discussion on the few-shot learning approaches that are explicitly designed for semantic segmentation of aerial imagery.
\subsection{Semantic Segmentation for Visual Scenes}
\indent Semantic segmentation forms a critical research area within computer vision, carrying substantial impact on interpreting visual scenes. Several works executed on traditional datasets \cite{mao2022bidirectional,kirillov2023segment} have relied heavily on conventional Convolutional Neural Networks (CNNs), yielding valuable but large-dataset-dependent methods \cite{yuan2021review,diakogiannis2020resunet}. They provide valuable insights but often struggle with time-efficiency due to their heavy reliance on extensive annotated datasets.
\subsection{Few-shot Learning and Large Vision Models}
\indent Few-shot learning, with its ability to generalize from limited data, has recently emerged as an effective approach for semantic segmentation \cite{wang2020generalizing,sung2018learning}. However, the scope of these works does not extend towards the incorporation of large vision models. In recent years, Large Vision Models, like GPT-3 \cite{dale2021gpt} and CLIP \cite{radford2021learning}, have sparked interest due to their impressive performance in numerous visual and text-based tasks, including SAM \cite{kirillov2023segment} due to its specialized segmentation capabilities.
\subsection{Few-Shot Learning in Semantic Segmentation of Remote Sensing Imagery}
\indent Efforts specifically centred around applying few-shot learning for the semantic segmentation of remote sensing imagery remain sparse \cite{mao2022bidirectional,wang2021dmml}. The studies by \cite{he2023accuracy,wu2023medical} have revealed promising gains in using SAM for semantic segmentation tasks. Still, their focal point is largely limited to medical imaging, despite the high potential applicability and relevance to remote sensing imagery. 
\section{Methodology}
Our methodology outlines the systematic integration of the Segment Anything Model (SAM), Prior Guided Metric Learning, and an innovative few-shot learning setup to effectively automate semantic segmentation within the setting of remote sensing imagery. To provide more insight, we denote the training dataset as $D = \{I, T\}$, support set as $D_s = \{I_s, T_s\}$, and query set as $D_q = \{I_q, T_q\}$, where $I = \{i^1, ... , i^n\}$ represents images, and $T = \{t^1, ... , t^n\}$ corresponds to their segmentation ground truth. The primary objective is to design a plug-and-play self-prompting module, enabling SAM to obtain the location and size information of the segmentation target, only necessitating a few labeled data, for instance $k$ images.\\
\begin{figure}[t]
\includegraphics[width=\textwidth]{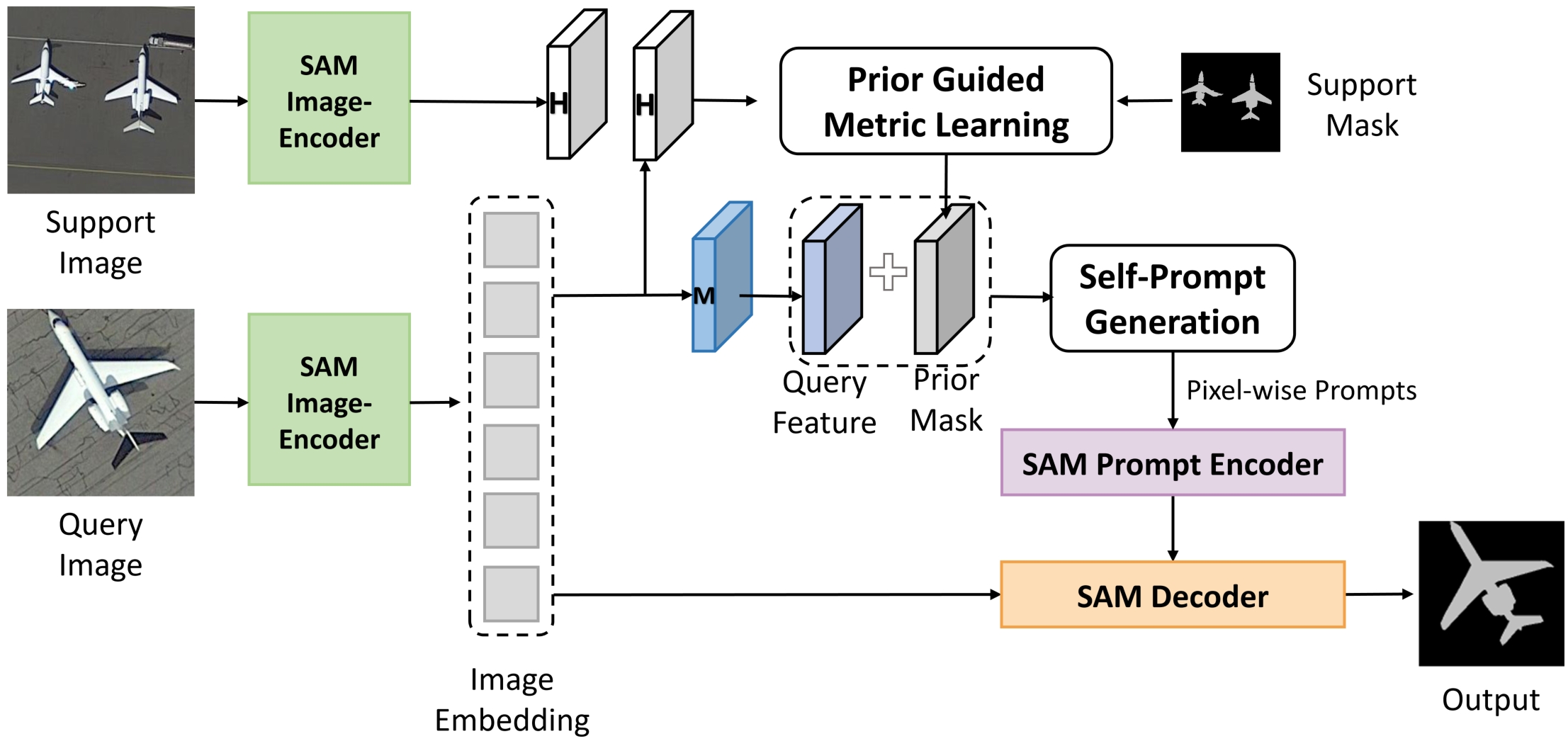}
\caption{The overview of our proposed Few-shot Semantic Segmentation framework based on Self-guided Large Vision Model (Few-shot SLVM). We utilize the pretrained Large Vision Model, SAM, to extract both high-level (H) and intermediate (M) semantic features from support and query images. The pre-trained high-level features with a support mask are transformed into prior mask utilizing cosine similarity measures. We take the query features and generated prior mask as input to produce coarse pixel-wise prompts for SAM. During the training process, the model is trained using the image embeddings from the SAM encoder and the resized ground truth label, while the cumbersome encoder, prompt encoder, and decoder parts of the SAM structure are kept frozen.} \label{overview}
\end{figure}
\subsection{Prior Guided Metric Learning}
Assuming an input image $X$ and the output segmentation mask $Y$, with the encoder function $E(\cdot)$ and decoder function $D(\cdot)$ of the Segment Anything Model (SAM), we can define the process of generating the mask for SAM as follows:
\begin{equation}
Y = D(E(X))
\end{equation}
where $E$ represents a pre-trained image embedding encoder and $D$ is a learned mask decoder, the Prior Guided Metric Learning Module is introduced to incorporate prior information with the prompt. Specifically, after passing through the powerful encoder of the large vision model SAM, we first perform the Hadamard product between the high-level support features $E_H(I_S)$ and the mask $M$. Subsequently, we use cosine similarity to calculate the pixel-wise association between the high-level query features $E_H(I_Q)$ and the mask-weighted support features, defined as:
\begin{equation}
P = cosine(E_H(I_Q), E_H(I_S)\odot M)
\end{equation}
By concatenating the intermediate query features $E_M(I_Q)$ with the pixel-level prior-guided information $P$, new query features are generated to effectively integrate the support information with the prior information, resulting in enhanced segmentation results.
\subsection{Automatic Prompt Learning}
In this step, we design a novel automatic prompt learning method that generates coarse pixel-wise prompts for SAM from the prior mask $Y_P$ to guide the segmentation prediction. The prompt indicators $W$ are derived from the prior mask $Y_P$, and the output mask Y is reformulated as:
\begin{equation}
Y = D(W \odot E(X), (1 - W) \odot E(Y_P))
\end{equation}
During the training process, both the cumbersome encoder and decoder of SAM are kept frozen, guiding it to focus on the area of interest through the continuous optimization of self-guided prompt embeddings.
\subsection{Few-Shot Learning Adaptation}
Our proposed Few-shot Self-guided Large Vision Model (SLVM) functions by learning from a limited set of support examples and extrapolating this learning to the query set. For this, we employ a cosine similarity loss function, defined as:
\begin{equation}
L = \frac{1}{N} \sum_{i=1}^N\left(1-\frac{Y_i \cdot T_i}{\left\|Y_i\right\| \cdot\left\|T_i\right\|}\right)
\end{equation}
Lastly, to enhance performance, we introduce a fine-tuning strategy. The training objective comprises both the Self-guidance loss $L_s$ and the Fine-tuning loss $L_f$, represented as:
\begin{equation}
L_{total} = \alpha L_s + \beta L_f
\end{equation}
The strategy operates with a two-fold phase that initially trains with self-guidance loss only and then fine-tunes using the total loss. Through this detailed walkthrough of our method, we lay out the blueprint of our Few-shot SLVM model’s capability to combine the power of SAM, few-shot learning, and prior metric learning for semantic segmentation in remote sensing images.
\section{Experiments}
\subsection{Datasets }
\textbf{DLRSD} dataset \cite{shao2018benchmark} employed in the experiments comprises 2100 high-resolution aerial images, with each image having dimensions of 256 × 256 pixels. These images encompass 17 distinct object classes, including airplanes, bare soil, buildings, cars, and various others. Each sample in the dataset is labeled with pixel-level annotations, providing detailed ground truth information for precise object segmentation. The dataset poses several challenges encountered in real-world scenarios, such as occlusion, shadows, and variations in terrain scales, making it a valuable resource for evaluating algorithms robustness. Similar to the methodology of Wang et al. \cite{wang2021dmml}, the DLRSD dataset is partitioned into four separate folds. The first three folds consist of four categories each, while the fourth fold includes five categories, namely sea, ship, tank, tree, and water. This partitioning allows for a more comprehensive evaluation and analysis of the proposed method's performance on different object classes and scenarios.
\subsection{Implement Details}
For the following experiments, we employ the SAM ViT-Huge model as our backbone. During the training phase, we utilize the PyTorch framework and trained end-to-end with the AdamW optimizer. We use a mini-batch size of 8 and set the initial learning rate to 0.00025. To decay the learning rate, we employ a Cosine Annealing scheduler and the momentum is set to 0.9. For data augmentation, we randomly perform flipping (vertically or horizontally) and rotation operations on the input images, with a resulting size of 256x256 pixels. All experiments are conducted on 2 NVIDIA GeForce RTX 3090 Ti GPUs, and the training process lasts for 1000 epochs. By utilizing appropriate pretrained backbones and carefully setting hyperparameters, we achieve optimal performance in terms of accuracy and efficiency in our experiments.
\subsection{Results}
To provide a comprehensive evaluation, we compared our proposed Few-shot SLVM with three other state-of-the-art few-shot segmentation methods evaluated on the DLRSD dataset, taking into account the 1-shot and 5-shot settings, and utilizing mean intersection over union (mIoU) as the evaluation metric.
\begin{table*}
\centering
\caption{Comparisons of few-shot segmentation performance between our proposed Few-shot SLVM and other methods under different splits on the DLRSD dataset.}
\begin{tabular}{ccccccccccc}
\toprule
\multicolumn{1}{l|}{\multirow{2}{*}{Methods}} & \multicolumn{5}{c|}{1-Shot} & \multicolumn{5}{c}{5-Shot}                                                                                       \\ \cmidrule{2-11} 
\multicolumn{1}{l|}{}                         & \multicolumn{1}{c}{Fold-0} & \multicolumn{1}{c}{Fold-1} & \multicolumn{1}{c}{Fold-2} & \multicolumn{1}{c|}{Fold-3} & \multicolumn{1}{c|}{Mean} & \multicolumn{1}{c}{Fold-0} & \multicolumn{1}{c}{Fold-1} & \multicolumn{1}{c}{Fold-2} & \multicolumn{1}{c|}{Fold-3} & \multicolumn{1}{c}{Mean} 
\\ \midrule
\multicolumn{1}{l|}{CANet} & 25.31 & 12.55 & 18.41 & \multicolumn{1}{l|}{26.66} &\multicolumn{1}{l|}{20.73} & 28.29 & 17.10 & 21.36 & \multicolumn{1}{l|}{29.45} & 24.05     \\
\multicolumn{1}{l|}{PANet} & 36.15 & 20.55 & 26.98 & \multicolumn{1}{l|}{38.41} &\multicolumn{1}{l|}{30.52}     & 40.85 & 23.61 & 35.87 & \multicolumn{1}{l|}{45.67} & 36.50 \\
\multicolumn{1}{l|}{DMML-Net} & 45.03 & 31.23 & 47.38 & \multicolumn{1}{l|}{\textbf{47.17}} &\multicolumn{1}{l|}{42.70}     & 57.23 & 39.86 & 56.62 & \multicolumn{1}{l|}{\textbf{62.60}} & 54.08 \\
\multicolumn{1}{l|}{Ours} & \textbf{50.70} & \textbf{34.15} & \textbf{50.47} & \multicolumn{1}{l|}{43.64} & \multicolumn{1}{l|}{\textbf{44.74}}     & \textbf{61.90} & \textbf{52.20} & \textbf{58.72} & \multicolumn{1}{l|}{60.05}       &\textbf{58.22}      \\ \bottomrule
\end{tabular}
\label{tab2}
\end{table*}

\noindent Comparative Analysis: As Table 1 depicts, Few-shot SLVM surpasses competing methodologies, underscoring its robustness and accuracy. The mean IoU at the 1-shot and 5-shot settings improved by 2.04\% and 4.14\% over the closest competing models, which demonstrates the model's superior segmentation quality.
\subsection{Ablation Study}
We performed an ablation study to understand the contribution of each component in our model. We focused on three main modules: (1) Automatic Prompt Learning (APL), (2) Prior Guided Metric Learning (PGML), and (3) Few-Shot Learning Adaptation (FSLA). Results from the ablation study:
\begin{table}[]
\begin{center}
\caption{ABLATION STUDY ON THE PROPOSED COMPONENTS OF OUR METHOD ON THE DLRSD DATASET UNDER ONE-SHOT SETTING.}
\begin{tabular}{cccccccc}
\toprule
\multicolumn{3}{c|}{\multirow{1}{*}{Methods}} & \multicolumn{1}{c}{\multirow{2}{*}{Fold-0}} & \multicolumn{1}{c}{\multirow{2}{*}{Fold-1}} & \multicolumn{1}{l}{\multirow{2}{*}{Fold-2}} & \multicolumn{1}{l|}{\multirow{2}{*}{Fold-3}} & \multicolumn{1}{c}{\multirow{2}{*}{Mean}} 
\\ \cmidrule{1-3} 
\multicolumn{1}{l}{\multirow{1}{*}{APL}} & \multicolumn{1}{l}{\multirow{1}{*}{PGML}} & \multicolumn{1}{l|}{\multirow{1}{*}{FSLA}}  & & & & \multicolumn{1}{l|}{} 
\\ \midrule
 &  & \multicolumn{1}{l|}{} & 39.02 & 21.19 & 38.70 & \multicolumn{1}{l|}{36.04} &\multicolumn{1}{l}{33.74} \\
\checkmark &  & \multicolumn{1}{l|}{} & 46.77 & 27.30 & 46.44 & \multicolumn{1}{l|}{41.85} &\multicolumn{1}{l}{40.59} \\
\checkmark & \checkmark & \multicolumn{1}{l|}{} & 50.66 & 31.35 & 45.03& \multicolumn{1}{l|}{42.16} &\multicolumn{1}{l}{42.30} \\
\checkmark & \checkmark & \multicolumn{1}{l|}{\checkmark} & \textbf{50.70} & \textbf{34.15} & \textbf{50.47} & \multicolumn{1}{l|}{\textbf{43.64}} & \multicolumn{1}{l}{\textbf{44.74}}      \\ \bottomrule
\end{tabular}
\end{center}
\label{tab3}
\end{table}

\noindent Ablation Study Insights: The stark performance drop in the absence of 'Automatic Prompt Learning' (Table 2) confirms its centrality in generating precise segmentation prompts, contributing to an overall 6.85\% increase in Mean IoU. Similarly, disabling 'Few-Shot Learning Adaptation' leads to performance degradation, highlighting its role in enhancing the model's adaptability to limited data scenarios.\\
\indent Module Effectiveness: 1. Automatic Prompt Learning: The value of the APL component is evident from the substantial uptick in the overall accuracy. Through the automation of prompt generation, it diminishes reliance on manual interventions. This hastens the segmentation workflow and either maintains or heightens the quality of the results. The difference in the mean IoU between the model with no components and the one with APL indicates a marked improvement of 6.85\%, underscoring the importance of APL. 2. Prior Guided Metric Learning (PGML): Delving into the performance with and without the PGML, the significance of this module comes to light. When we compare the model with only APL to the one equipped with both APL and PGML, there's a modest increase in Mean IoU from 40.59\% to 42.30\%. This suggests that PGML refines the feature representations and contributes to better distance metrics. By incorporating prior knowledge, PGML offers a more informed and directed approach to metric learning, thus making the model more resilient and effective in differentiating between classes. 3. Few-Shot Learning Adaptation: The FSLA component bolsters the model's ability to learn from limited data. This trait is paramount, especially in remote sensing scenarios where access to abundant labeled data might be constrained. Its role is apparent in the enhancement of the model's generalization capabilities, especially on classes that are either unseen or minimally represented. The final row of the table demonstrates that integrating FSLA boosts the mean IoU to 44.74\%, the best among the configurations.

\section{Conclusion}
In conclusion, this research emphasizes the advancements brought forth by the Few-Shot Self-guided Large Vision Model (Few-shot SLVM) in few-shot semantic segmentation for remote sensing imagery. The Few-shot SLVM integrates three pivotal modules: Automatic Prompt Learning (APL), Prior Guided Metric Learning (PGML), and Few-Shot Learning Adaptation (FSLA). APL streamlines the segmentation process, reducing manual input and enhancing accuracy. PGML optimizes feature representation, refining class differentiation and subsequently boosting the mean IoU. FSLA excels in limited data scenarios, improving model generalization for rare or unseen classes. Collectively, these components elevate the Few-shot SLVM to deliver state-of-the-art results in few-shot remote sensing segmentation. Future endeavors will focus on refining these modules and extending their applicability across broader datasets for robust model validation.

%
%
%
%
\bibliographystyle{splncs04}
\bibliography{main}
\end{document}